# Building Odia Shallow Parser

**Pruthwik Mishra  Dipti Misra Sharma**
Language Technologies Research Centre
IIIT Hyderabad
pruthwik.mishra@research.iiit.ac.in
dipti@iiit.ac.in

## Abstract

Shallow parsing is an essential task for many NLP applications like machine translation, summarization, sentiment analysis, aspect identification and many more. Quality annotated corpora is critical for building accurate shallow parsers. Many Indian languages are resource poor with respect to the availability of corpora in general. So, this paper is an attempt towards creating quality corpora for shallow parsers. The contribution of this paper is two folds: creation pos and chunk annotated corpora for Odia and development of baseline systems for pos tagging and chunking in Odia.

## 1 Introduction

Odia or Oriya is an Indian language from the Indo-Aryan branch and is a part of the larger Indo-European language family. It is the native language of Odisha [1] state spoken by around 50 million people. There are many Indian Languages which are unexplored due to the unavailability of annotated corpora. Odia is one of them. The motivation behind this work is to create lexical resources for the Odia language and improve the accuracy of the Part Of Speech Tagger and Chunker.

Shallow Parsing involves Tokenization, Part Of Speech Tagging(POS) and Chunking that are explained in the section 2. We present an approach for building a shallow parser for Odia. We used conditional random fields(CRF) algorithm for POS Tagging and Chunking. We report 81% accuracy for the Odia Shallow Parser on ILCI [2] data-set from Agriculture domain. We implemented the Odia POS Tagger and the Odia Chunker in this work.

## 2 Shallow Parser Pipeline

Shallow Parsing acts as an initial step to full parsing. We follow a pipeline approach for creating the shallow parser. Tokenization, Part-Of-Speech(POS) (Bharati et al., 2007) Tagging and Chunking are the different modules arranged in the shallow parser pipeline. The Shallow Parser's performance is dependent on the accuracy of each module.

### 2.1 Tokenization

The first step in a shallow parser is Tokenization. The in-house tokenizer is created as a part of ILMT [3] project is used for this work [4]. There remains ambiguities with respect to the sentence boundary identification. We resolve those through manual analysis. Other issues are handled by hand-written rules.

### 2.2 POS Tagging

The next step is POS Tagging. For automatic POS tagging, the state-of-the-art POS tagger uses large POS annotated data sets and learns the appropriate class labels for words depending on various hand annotated features. Odia is a fusional language. Prefixes and the word ending suffixes encode a lot information about the category of the word. We leverage these features for building a robust POS Tagger and in turn the shallow parser.

---

[1] https://en.wikipedia.org/wiki/Odisha
[2] http://hnk.ffzg.hr/bibl/ltc2013/book/papers/LRT2-4.pdf

[3] IL-ILMT is project sponsored by MC&IT, Govt. of India Reference No: 11(10)/2006-HCC(TDIL)
[4] No Accuracy has been reported for the tokenizer

## 2.3 Chunking

The next step in the shallow parser is Chunking (Abney, 1992). Chunking is the process of identifying non-recursive chunks present in a sentence. Each chunk contains a head and some modifiers modifying the head. Chunking is highly reliant on POS Tagging. So, it becomes crucial to solve the ambiguities related to POS Tagging for the overall accuracy of the Chunker, as a result shallow parser.

The paper is organized as follows. In the section 2, we describe the Background of Shallow Parsing. The next section discusses the previous works done in Odia shallow parsing. The Corpus Details are provided in Section 4. The annotation procedure is described in section 5. We describe the Approach and various tools used in section 6. Section 7 presents the Experimental Results and subsequently, we present the error analysis. The future work is discussed in the concluding section. The POS tagger is available at https://github.com/Pruthwik/Odia-POS-Tagger and Chunker at https://github.com/Pruthwik/Odia-Chunker

## 3 Related Work

Compared to other Indian languages, Odia is a resource poor language. There have been attempts to build shallow parsers for Odia using very limited annotated data. There state-of-the-art POS Tagger for Odia using SVM (Das et al., 2015) classifier reported around 82% accuracy on a tag-set containing only 5 tags [5]. We report the results using the tagset designed for Indian Languages (IIIT tagset) (IIIT-tagset, 2007) which is fine-grained. The number of pos tags and chunk tags in the ILMT tagset is 27 and 11 respectively.

## 4 Corpus Details

We collected publicly available the raw ILCI (Choudhary and Jha, 2014) corpus from Agriculture domain for Odia. We manually annotated 700 sentences containing 8.5K tokens following Odia grammar guidelines and evaluated the performance of the POS Tagger and the Chunker on the annotated corpus.

---
[5] Verb, Noun, Pronoun, Adverb and Adjective, There was no information about the corpus

| Type  | #Sentences | #Tokens |
|-------|------------|---------|
| All   | 700        | 8.5K    |
| Train | 500        | 6.2K    |
| Test  | 200        | 2.3K    |

Table 1: Corpus Details

The details of the corpus used in this work are presented in Table 1. We split the annotated dataset into train and test datasets.

## 5 Annotation Procedure

POS tagging and chunking involves annotation of each token appearing in a sentence. Without the help of any annotation tool, it becomes a herculian task. So, we used Sanchay (Singh and Ambati, 2010) tool for facilitating the annotation task. It is easily configurable and customizable. Custom POS tags and chunk tags can be easily added in the tool. The annotations are saved in Shakti Standard format (Bharati et al., 2007). A sample of annotation under Sanchay is shown in the figure 1. Annotation guidelines for the tasks were prepared. 2 annotators are involved in the task who are post graduate students. The inter annotators' agreement is found to be 0.91 and 0.87 for POS tagging and chunking respectively in terms of Fleiss' Kappa. This shows perfect agreement between the annotators. Some pos and chunk tags from the ILMT tagset are shown in tables 2 and 3.

| Tag  | Meaning                    |
|------|----------------------------|
| NN   | Common Noun                |
| NNP  | Proper Noun                |
| NST  | Locative and Spatial Noun  |
| PRP  | Pronoun                    |
| PSP  | Postposition               |
| JJ   | Adjective                  |
| RB   | Adverb                     |
| CC   | Conjunction                |
| VM   | Main Verb                  |
| VAUX | Auxiliary Verb             |

Table 2: Chunks Tags

## 6 Approach

We used conditional random fields(CRF)[6] for POS Tagging and Chunking tasks because CRF (Agrawal, 2007) achieves better results than other generative models and Maximum Entropy models for sequential labeling tasks. For POS Tagging, Chunking we designed different sets of features.

---
[6] https://taku910.github.io/crfpp/

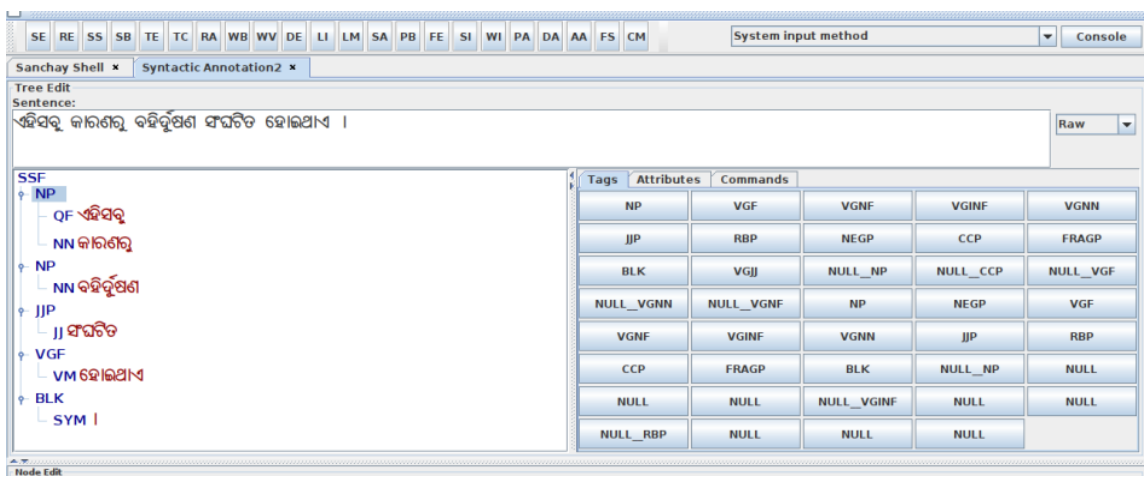

Figure 1: Annotation in Sanchay

| Tag | Meaning |
|---|---|
| NP | Noun Phrase |
| JJP | Adjectival Phrase |
| RBP | Adverbial Phrase |
| VGF | Finite Verb Phrase |
| VGNF | Non-Finite Verb Phrase |

Table 3: Chunks Tags

## 6.1 Feature Selection for POS Tagging

As POS Tagging is a sequence labeling task, certain features need to be captured in classifying the words and assigning them appropriate tags. In Odia, prefixes and suffixes provide a lot of information about the category of the word.

We designed the following features for POS Tagging for Odia.

- Current Word

- The prefixes from length 1 to maximum length m

- The suffixes from length 1 to maximum length n

- Length of the word

- Token Context Window size of s $w_{-i} \ldots w_{-2} w_{-1} w_0 w_{+1} \ldots w_{+i}$

where $w_0 \rightarrow$ current token, $w_{-1} \rightarrow$ previous token, $w_{+1} \rightarrow$ next token, and so on

$m = 4, n = 7, s = 1$ is found to be the best parameters for Odia from experiments with different values of $m, n, s$.

### 6.1.1 Features for Odia

$m = 4, n = 7, s = 1$ is found to be the best parameters for Odia from experiments with different values of $m, n, s$.

## 6.2 Chunk Features

For Chunking the below features have been used:

- Current Word

- Current POS Tag

- Token Context Window of Size $s_1$

- Context Window for POS Tags of Size $s_2$

$s_1 = 1, s_2 = 1$ is found to be the best parameters for Odia chunking.

## 7 Results & Error Analysis

The F1 scores of POS tagging and chunking are 0.81 and 0.83 respectively. The shallow parsing pipeline consists of POS tagging and chunking in sequence. The output of the POS tagger is given to the chunker as a feature while the chunker is trained on gold pos tags as a feature. The sources of major errors are listed as follows:

| Model | P | R | F1 |
|---|---|---|---|
| POS Tagging | 0.82 | 0.80 | 0.81 |
| Chunking | 0.83 | 0.83 | 0.83 |
| Shallow Parsing | 0.81 | 0.82 | 0.81 |

Table 4: Results for POS Tagging, Chunking, and Shallow Parsing in Odia

- Ambiguities between Nouns and Adjective
- Ambiguities between Nouns and Proper Nouns
- Ambiguities related to Compounds
- Ambiguities between Verbs and Auxiliary Verbs
- Ambiguities in Question words, Quantifiers and Demonstratives

Example of ambiguities in Odia:

ଭଲ ପିଲାଙ୍କୁ ଖାଦ୍ୟ ଦିଅ
good boy-accusative food give
Give food to the good boy

ଭଲଙ୍କୁ ଖାଦ୍ୟ ଦିଅ
good-accusative food give
Give food to the good one

ଭଲ(good) is used as an adjective in first sentence while it is used as a noun in the second one. But the POS tagger erroneously tags the second token as adjective. In the second example, the suffix କୁ acts as the case marker(accusative) and this suffix disambiguates the noun from the adjective used in the first example. This can be used as a feature to improve the POS tagger.

Similarly the word ଏପରି(gloss - in this way) is ambiguous between pronoun(PRP) and demonstrative(DEM). The presence of a noun in the next word will disambiguate this, therefore we need to consider the context of next word.

Still the errors pertaining to the tag RDP persists. As the RDP (reduplication) tag depends on the previous word, the previous token may be tagged as a noun (NN), adjective (JJ), verb (VM) and other categories. The reduplicated word exhibits the same features as the previous word, so it is most likely to be tagged as the previous word. Ambiguities between Question Words (WQ) and Quantifiers (QF) exist in Odia between the words କିପରି(how) and ଏପରି(this way).

## 8 Conclusion & Future Work

In this paper, we created POS tag and chunk annotated corpora in Odia. We developed baseline models using CRF as well. We hope that this will act as an impetus for corpora creation and development of better models in Odia. The performance of Odia shallow parser is impacted by the smaller training data size. As a future work, we intend to implement bootstrapping techniques to assist in increasing the size of annotated corpora for Odia. It will be interesting to observe how the shallow parser performs on BIS(Bureau of Indian Standards) tagset. We will explore LSTM and BERT based fine tuning for shallow parsing as well.